\documentclass{article}
\usepackage{spconf,amsmath,graphicx,hyperref}

\usepackage[utf8]{inputenc} 
\usepackage[T1]{fontenc}    
\usepackage{hyperref}       
\usepackage{url}            
\usepackage{booktabs}       
\usepackage{amsfonts}       
\usepackage{nicefrac}       
\usepackage{microtype}      
\usepackage{xcolor}         
\usepackage{adjustbox}
\usepackage{subcaption}
\usepackage{wrapfig} 
\usepackage{paralist}
\usepackage{makecell}
\usepackage{enumitem}

\usepackage{amsmath,epstopdf,amssymb,color,url,multirow,multicol,verbatim,threeparttable,diagbox,makecell,booktabs,color,colortbl}  
\definecolor{myred}{RGB}{211, 47, 47}    
\definecolor{myteal}{RGB}{0, 121, 107}   
\definecolor{mygray}{rgb}{0.9, 0.95, 1}
\definecolor{rowdarkgreen}{RGB}{200, 230, 201} 
\definecolor{accentorange}{RGB}{255, 204, 128}
\definecolor{accentblue}{RGB}{144, 202, 249}
\usepackage{graphicx}
\usepackage{tcolorbox}
\usepackage{courier}

\title{Brought a Gun to a Knife fight: Modern VFM Baselines Outgun Specialized Detectors on In-the-Wild AI Image Detection
}
\name{Yue Zhou$^{1}$, Xinan He$^{1,4}$, Kaiqing Lin$^{1}$, 
      Bing Fan$^{2}$, Feng Ding$^{4}$, Jinhua Zeng$^{3}$, 
      Bin Li$^{1}$\sthanks{Corresponding author.}}

\address{$^{1}$ Guangdong Provincial Key Laboratory of Intelligent Information Processing,\\
         Shenzhen Key Laboratory of Media Security,\\ and SZU\textendash AFS Joint Innovation Center for AI Technology, Shenzhen University \\
         $^{2}$ University of North Texas \\ 
         $^{3}$ Academy of Forensic Science\\
         $^{4}$ Nanchang University\\}
\begin{document}
%
\maketitle
\begin{abstract}
While specialized detectors for AI-generated images excel on curated benchmarks, they fail catastrophically in real-world scenarios, as evidenced by their critically high false-negative rates on `in-the-wild' benchmarks. Instead of crafting another specialized `knife' for this problem, we bring a `gun' to the fight: a simple linear classifier on a latest Vision Foundation Model (VFM). Trained on identical data, this baseline decisively `outguns' bespoke detectors, boosting in-the-wild accuracy by a striking margin of over 20\%.

Our analysis pinpoints the source of the VFM's `firepower': First, by probing text-image similarities, we find that recent VLMs (e.g., Perception Encoder, Meta CLIP-2) have learned to align synthetic images with forgery-related concepts (e.g., `AI-generated'), unlike their previous versions. Second, we speculate that this is due to data exposure, as performance plummets on a \textbf{verifiably unseen dataset} composed of post-cutoff synthetic images and private photographs. Our findings yield two critical insights: 1) For the real-world `gunfight' of AI-generated image detection, the raw `firepower' of an updated VFM is far more effective than the `craftsmanship' of a static detector. 2) True generalization evaluation requires test data to be independent of the model's entire training history, including pre-training.
\end{abstract}

\section{Introduction}
\label{sec:intro}
The field of Artificial Intelligence Generated Images (AIGI) has seen explosive growth in recent years, with a particular focus on the creation of highly realistic synthetic images through advanced generative models. These synthetic images have significantly fueled the proliferation of misinformation, creating serious threats to both societal security and individual privacy. 

\begin{figure}[t!]
  \centering
  \includegraphics[width=0.9\linewidth]{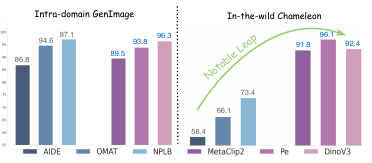}
  \vspace{-10pt}
  \caption{Comparison of detection performance between modern VFMs and state-of-the-art forensics-specialized detectors on the GenImage\cite{zhu2023genimage} and Chameleon\cite{yan2024sanity}.}
  \label{fig:1}
  \vspace{-15pt}
\end{figure}

As a result, a central challenge for AIGI detection is building models with a strong generalization ability, allowing them to effectively identify and verify images generated by a variety of unknown methods. To address this, researchers have developed various forensics-specialized detectors, such as incorporating additional augmented samples (e.g., full or partial image reconstructions) \cite{ojha2023towards,chen2024drct,ding2025fairadapter} or designing improved training strategies \cite{mu2025no, lin2025seeing, he2025vlforgery}. These approaches have already pushed the performance on recent AIGI benchmarks to impressive levels. Consequently, the research focus has begun to shift toward the more challenging in-the-wild datasets. However, such real-world scenarios remain highly challenging, as most forensic methods fail to reach satisfactory performance, particularly on the Chameleon \cite{yan2024sanity}, as illustrated in Fig.~\ref{fig:1}.

In this work, we show that a surprisingly simple baseline: a linear classifier using a modern VFM as a feature extractor, training the same examples without any augmentation, already outperformed even the most advanced forensic detectors on in-the-wild datasets, improving accuracy by over 20\%. In particular, a new generation of foundation models, spanning both Vision-Language (e.g., PE\cite{bolya2025perception}, Meta CLIP-2\cite{chuang2025metaclip}) and self-supervised architectures (e.g., DINOv3\cite{simeoni2025dinov3}), though not being specifically designed for forensics, exhibit strikingly superior performance compared to both earlier VFMs and forensics-specific detectors. This raises a critical question: does their success arise from genuinely stronger representations, or from inadvertent exposure to synthetic data during pre-training?

To investigate this issue, we conduct two analyses. We first test all VFMs on a \textbf{verifiably unseen dataset} composed of post-cutoff synthetic media and private, personally captured photographs. As hypothesized, the performance of most modern VFMs drops significantly, revealing their reliance on prior data exposure. Building on this, our second analysis probes the underlying mechanism for Vision-Language Models. We find they have learned to align synthetic images with \textbf{forgery-related concepts} (e.g., `AI-generated'), providing a semantic explanation for their effectiveness. This 'built-in' knowledge, learned from web-scale data, allows these general-purpose models to decisively outgun specialist detectors explicitly designed for the forensic task.

In summary, our main contributions are:
\begin{itemize}[itemsep=4pt, topsep=4pt, parsep=0pt, partopsep=0pt]
    \item We establish the superiority of simple, modern VFM baselines over specialized detectors for in-the-wild scenarios. We argue that for \textit{real-world effectiveness}, the best strategy is to leverage the most up-to-date pre-trained knowledge, even if it stems from data exposure.
    \item By identifying data exposure as the primary source of this success, we reveal a fundamental flaw in static benchmarks and thus advocate for a crucial shift towards dynamic, continuously updated evaluation protocols. This will ensure they remain verifiably unseen, providing a more accurate and future-proof measure of a model's ability to handle genuinely novel threats.
\end{itemize}

\vspace{-4mm}
\section{Evaluation}
\label{sec:methology}
\vspace{-1mm}

\subsection{Experimental Setup}

\noindent
\textbf{Models.} We evaluate a diverse set of Vision Foundation Models (VFMs), categorized into three groups: 1) \textbf{Modern VFMs} released after early 2025 (Meta CLIP-2~\cite{chuang2025metaclip}, Pe~\cite{bolya2025perception}, SigLIP-2~\cite{tschannen2025siglip}); 2) their direct \textbf{Predecessors} (CLIP~\cite{radford2021learning}, Meta CLIP~\cite{xu2024altogether}, SigLIP~\cite{zhai2023sigmoid}); and 3) \textbf{Self-supervised Models} (DINOv3~\cite{simeoni2025dinov3} and DINOv2~\cite{oquab2023dinov2}). For all VFMs, we use their largest, officially released weights.

\noindent
\textbf{Evaluation Datasets.} Our evaluation focuses on established in-the-wild datasets, each designed to capture different facets of real-world scenarios. We primarily use four benchmarks which can be categorized by their collection sources:

\begin{itemize}[itemsep=5pt, topsep=5pt, parsep=0pt, partopsep=0pt]
    \item \textbf{Social Media Sourced:} Datasets like \textbf{WildRF}~\cite{cavia2024real} and \textbf{SocialRF}~\cite{li2025artificial} are collected from broad social media platforms such as X (formerly Twitter), Facebook, and Reddit. Their collection process typically involves querying common hashtags associated with authentic content (e.g., `\#photography`, `\#nofilter`) and AI-generated content (e.g., `\#AIart`, `\#midjourney`).

    \item \textbf{AI Art Community Sourced:} In contrast, datasets like \textbf{Chameleon}~\cite{yan2024sanity} and \textbf{CommunityAI}~\cite{li2025artificial} are curated from specialized AI art communities like ArtStation and Civitai. These benchmarks prioritize high-realism, high-quality images from diverse models, often reflecting the cutting edge of generative art.
\end{itemize}

\noindent
\textbf{Training and Evaluation Protocol.} For each VFM, we train a single linear classifier on its frozen `[CLS]' token features, using the GenImage~\cite{zhu2023genimage} (SD v1.4 subset) \textbf{without any augmentations}. The resulting classifiers are then evaluated on all in-the-wild datasets. For comparison against SOTA forensic detectors, we report their performance directly from the original publications, as they often rely on specialized training strategies incompatible with our unified setup.

\begin{table}[t!]
\centering
\captionof{table}{Comparison results on GenImage and Chameleon.}\vspace{-10pt}
\label{tab:compare-genimagechameleon}
\begin{adjustbox}{width=0.8\linewidth} 
  \begin{tabular}{l|c|ccc}
  \toprule
  \multirow{2}{*}{Method} & GenImage & \multicolumn{3}{c}{Chameleon} \\
  \cmidrule(r){2-2} \cmidrule(r){3-5}
   & Avg. & R.Acc & F.Acc & Avg. \\
  \midrule
  CNNSpot~\cite{wang2020cnn}       & 64.2 & 99.7 & 0.08 & 56.9 \textsubscript{\textcolor{myred}{(-7.3)}} \\
  FreqNet~\cite{tan2024frequency}   & 86.8 & 87.1 & 13.7 & 55.6 \textsubscript{\textcolor{myred}{(-31.2)}} \\
  GramNet~\cite{liu2020global}      & 69.9 & 99.6 & 4.76 & 58.9 \textsubscript{\textcolor{myred}{(-11.0)}} \\
  UnivFD~\cite{ojha2023towards}      & 88.8 & 97.8 & 3.18 & 57.2 \textsubscript{\textcolor{myred}{(-31.6)}} \\
  NPR~\cite{cozzolino2024zero}         & 88.6 & 98.7 & 2.20 & 57.3 \textsubscript{\textcolor{myred}{(-31.3)}} \\
  AIDE~\cite{yan2024sanity}         & 86.8 & 98.5 & 5.04 & 58.4 \textsubscript{\textcolor{myred}{(-28.4)}} \\
  PPL~\cite{yang2025all}           & \textbf{97.2} & 68.1 & 64.7 & 66.6 \textsubscript{\textcolor{myred}{(-30.6)}} \\
  OMAT~\cite{zhou2025breaking}        & 94.6 & 90.2 & 33.9 & 66.1 \textsubscript{\textcolor{myred}{(-28.5)}} \\
  NPLB~\cite{mu2025no}          & \underline{97.1} & 79.4 & 65.5 & 73.4 \textsubscript{\textcolor{myred}{(-23.7)}} \\
  \midrule
  \rowcolor{mygray}
  CLIP \textsubscript{(2021.2.26)}       & 84.3 & 70.7 & 35.5 & 55.6 \textsubscript{\textcolor{myred}{(-28.7)}} \\
  \rowcolor{mygray}
  Siglip \textsubscript{(2023.3.27)}     & 88.6 & 70.8 & 49.3 & 61.6 \textsubscript{\textcolor{myred}{(-27.0)}} \\
  \rowcolor{mygray}
  Siglip2 \textsubscript{(2025.2.21)}    & 92.0 & 88.9 & 77.0 & 83.8 \textsubscript{\textcolor{myred}{(-8.2)}} \\
  \rowcolor{mygray}
  \textcolor{blue}{MetaCLIP} \textsubscript{\textcolor{blue}{(2023.9.28)}} & 76.6 & 37.6 & 91.5 & 60.7 \textsubscript{\textcolor{myred}{(-15.9)}} \\
  \rowcolor{mygray}
  \textcolor{blue}{MetaCLIP-2} \textsubscript{\textcolor{blue}{(2025.7.29)}} & 89.5 & 84.8 & 87.9 & 91.8 \textsubscript{\textcolor{myteal}{(+2.3)}} \\
  \rowcolor{mygray}
  \textcolor{blue}{PE} \textsubscript{\textcolor{blue}{(2025.4.17)}}       & 93.8 & 97.0 & 94.8 & \textbf{96.1} \textsubscript{\textcolor{myteal}{(+2.3)}} \\
  \rowcolor{mygray}
  \textcolor{blue}{DINOv2} \textsubscript{\textcolor{blue}{(2023.10.26)}}    & 85.3 & 62.9 & 58.1 & 60.8 \textsubscript{\textcolor{myred}{(-24.5)}} \\
  \rowcolor{mygray}
  \textcolor{blue}{DINOv3} \textsubscript{\textcolor{blue}{(2025.8.14)}}    & 96.3 & 93.7 & 90.5 & \underline{92.4} \textsubscript{\textcolor{myred}{(-3.9)}} \\
  \bottomrule
  \end{tabular}
\end{adjustbox}
\vspace{-15pt}
\end{table}

\vspace{-5mm}
\subsection{Performance on Curated Benchmarks vs. Chameleon}
\vspace{-1mm}

We first evaluate a wide range of specialized forensic detectors and our VFM baselines on both the curated GenImage benchmark and the challenging in-the-wild Chameleon dataset. The results, presented in Table~\ref{tab:compare-genimagechameleon}, reveal a stark performance dichotomy. 

On the GenImage benchmark, many recent specialized detectors like \textbf{PPL} (97.2\%) and \textbf{NPLB} (97.1\%) achieve impressive accuracy, often leveraging complex training strategies. However, this success is brittle and fails to translate to the more realistic Chameleon dataset. A dramatic performance collapse is observed across most specialized methods; traditional detectors like \textbf{CNNSpot} and \textbf{UnivFD} fail entirely, with Fake Accuracy (F.Acc) dropping to near zero. This reveals a critical disconnect between benchmark performance and the generalization required for real-world application.

In stark contrast, modern VFM baselines demonstrate exceptional effectiveness. Despite their simple training, models like \textbf{PE}, \textbf{Meta CLIP-2}, and \textbf{DINOv3} surpass 90\% accuracy on Chameleon, decisively outperforming the entire class of specialized detectors. This suggests that for practical, in-the-wild scenarios, the general-purpose representations of modern VFMs are fundamentally superior to task-specific designs.




\vspace{-3mm}
\subsection{Validation on Additional In-the-Wild Datasets}
\vspace{-1mm}

To further validate these findings, we selected several additional representative in-the-wild datasets (i.e., WildRF, CommunityAI and SocialRF) and conducted the same comparative experiments, as shown in Tab.~\ref{tab:compare-wildrf} and Tab.~\ref{tab:compare-AIGIbenchmark}. Nearly all forensic detectors lost their detection capabilities on WildRF, with only a few exceptions achieving good performance. However, all vision encoders surpassed a 70\% accuracy rate, and DINOv3 even reached an impressive 96.4\%. Similarly, Tab.~\ref{tab:compare-AIGIbenchmark} shows that all forensic-specialized detectors failed to perform on these datasets, while most vision encoders maintained excellent detection capabilities. For instance, Perception achieved a 96\% accuracy on the CommunityAI dataset, and DINOv3 reached 93\% on SocialRF.

\begin{table}[tb!]
\centering
\captionof{table}{Comparison results on WildRF.}
\vspace{-10pt}
\label{tab:compare-wildrf}
\begin{adjustbox}{width=0.8\linewidth}
  \begin{tabular}{l|cccc}
  \toprule
  Method &  Facebook & Reddit & Twitter & Avg. \\
  \midrule
  NPR \textsubscript{\textcolor{blue}{(ECCV'24)}} \cite{cozzolino2024zero} &  78.1 & 61.0 & 51.3 &63.5\\
  UnivFD \textsubscript{\textcolor{blue}{(CVPR'23)}} \cite{ojha2023towards} & 49.1 & 60.2 & 56.5 & 55.3 \\
  FatFormer \textsubscript{\textcolor{blue}{(CVPR'24)}} \cite{liu2024forgery} & 54.1& 68.1 & 54.4 & 58.9  \\
  SAFE \textsubscript{\textcolor{blue}{(KDD'25)}} \cite{li2025improving}  & 50.9& 74.1 & 37.5 & 57.2 \\
  C2P-CLIP \textsubscript{\textcolor{blue}{(AAAI'25)}} \cite{tan2025c2p}  & 54.4& 68.4 & 55.9 & 59.6\\ 
  AIDE \textsubscript{\textcolor{blue}{(ICLR'25)}} \cite{yan2024sanity}  & 57.8& 71.5 & 45.8 & 58.4 \\
  DRCT \textsubscript{\textcolor{blue}{(ICML'24)}} \cite{chen2024drct}  &  46.6 & 53.1 & 55.2 & 50.6\\ 
  AlignedForensics \textsubscript{\textcolor{blue}{(ICLR'25)}}\cite{rajanaligned} &89.4 & 69.1 & 81.8 & 80.1\\

  \rowcolor{mygray}
  CLIP \textsubscript{(2021.2.26)}&77.2&72.3&68.9&72.8\\
  \rowcolor{mygray}
  Siglip \textsubscript{(2023.3.27)}&58.4&66.5&72.2&65.7\\
  \rowcolor{mygray}
  Siglip2 \textsubscript{(2025.2.21)}&79.9&84.7&89.6&84.7 \\
  \rowcolor{mygray}
  \textcolor{blue}{Meta CLIP} \textsubscript{\textcolor{blue}{(2023.9.28)}}&63.8&74.1&77.9&71.9\\
  \rowcolor{mygray}
  \textcolor{blue}{Meta CLIP-2} \textsubscript{\textcolor{blue}{(2025.7.29)}} & 70.9 & 76.7 & 81.2 & 76.3\\
  \rowcolor{mygray}
  \textcolor{blue}{PE} \textsubscript{\textcolor{blue}{(2025.4.17)}}&81.6&84.9&88.8&85.1\\
  \rowcolor{mygray}
  \textcolor{blue}{DINOv2} \textsubscript{\textcolor{blue}{(2023.10.26)}}& 71.3 &71.0&71.3&71.2\\
  \rowcolor{mygray}
  \textcolor{blue}{DINOv3} \textsubscript{\textcolor{blue}{(2025.8.14)}} & 94.7 & 96.7 & 97.9 & \textbf{96.4} \\

  \bottomrule
  \end{tabular}
  \end{adjustbox}
  \vspace{-5pt}
\end{table}

\begin{table}[tb!]
\centering
\captionof{table}{Comparison results on SocialRF and CommunityAI.}
\vspace{-8pt}
\label{tab:compare-AIGIbenchmark}
\begin{adjustbox}{width=0.8\linewidth}
  \begin{tabular}{l|ccc|ccc}
  \toprule
  \multirow{2}{*}{Method} & \multicolumn{3}{c}{SocialRF}  & \multicolumn{3}{c}{CommunityAI} \\ \cmidrule(r){2-4}\cmidrule(r){5-7} 
   & R.Acc & F.Acc &Avg. &  R.Acc & F.Acc & Avg. \\
  \midrule
  CNNspot \textsubscript{\textcolor{blue}{(CVPR'20)}} \cite{wang2020cnn}& 94.7 & 7.5&51.1&97.3& 5.3&51.3\\ 
  GramNet \textsubscript{\textcolor{blue}{(CVPR'20)}} \cite{liu2020global}& 92.6 & 11.5 &52.1& 99.0 & 6.2&52.6\\
  UnivFD \textsubscript{\textcolor{blue}{(CVPR'23)}} \cite{ojha2023towards}& 53.3 & 55.5 &54.4& 82.8 & 51.2&67.0\\
  FreqNet \textsubscript{\textcolor{blue}{(AAAI'24)}} \cite{tan2024frequency}& 68.5 & 39.3 &53.9& 98.9 & 12.2&53.6\\
  NPR \textsubscript{\textcolor{blue}{(ECCV'24)}} \cite{cozzolino2024zero}& 96.3 & 21.9 &59.1& 99.9 & 8.2&54.05\\
  AIDE \textsubscript{\textcolor{blue}{(ICLR'25)}} \cite{yan2024sanity}& 97.2 & 18.4 &57.8& 99.0 & 9.3&54.15\\
  SAFE \textsubscript{\textcolor{blue}{(KDD'25)}} \cite{li2025improving}& 99.6 & 16.4 &58.0& 100.0 & 8.5&54.25\\
\midrule

  \rowcolor{mygray}
  CLIP \textsubscript{(2021.2.26)}&41.0&77.3&59.1&69.8&42.4&56.1\\
  \rowcolor{mygray}
  Siglip \textsubscript{(2023.3.27)}&47.3&62.3&54.8&49.7&55.2&52.4\\
  \rowcolor{mygray}
  Siglip2 \textsubscript{(2025.2.21)}& 64.8&90.2&\underline{77.5}&83.1&86.3&84.7 \\
  \rowcolor{mygray}
  \textcolor{blue}{Meta CLIP} \textsubscript{\textcolor{blue}{(2023.9.28)}}&40.5&86.4&63.5&35.5&93.4&64.5\\
  \rowcolor{mygray}
  \textcolor{blue}{Meta CLIP-2} \textsubscript{\textcolor{blue}{(2025.7.29)}}&41.6& 92.4&67.0&91.7&94.9&93.3  \\
  \rowcolor{mygray}
  \textcolor{blue}{PE} \textsubscript{\textcolor{blue}{(2025.4.17)}}&52.1 & 98.1 &75.1& 96.6 & 97.5&\textbf{97.1}\\
  \rowcolor{mygray}
  \textcolor{blue}{DINOv2} \textsubscript{\textcolor{blue}{(2023.10.26)}} & 60.3 & 69.5&64.9& 60.6 & 56.2 &58.4\\
  \rowcolor{mygray}
  \textcolor{blue}{DINOv3} \textsubscript{\textcolor{blue}{(2025.8.14)}} & 88.7 & 98.0&\textbf{93.3}& 95.6 & 95.0 &\underline{95.3} \\
  \bottomrule
  \end{tabular}
  \end{adjustbox}
  \vspace{-10pt}
\end{table}

\vspace{-2mm}
\section{Analysis}
\vspace{-2mm}
\subsection{Testing for Data Exposure with Verifiably Unseen Data}
\vspace{-2mm}

Our primary speculation is that the superior in-the-wild performance of modern VFMs is a direct result of \textbf{exposure to contemporary synthetic media during pre-training}, rather than an intrinsic leap in generalization. To rigorously test this, we constructed a new dataset designed to be a decisive blind test, where both real and fake images are verifiably unseen. Its synthetic and real portions are both designed to be verifiably unseen. The former comprises \textit{r/midjourney} images posted after all model training cut-off date, while the latter consists of \textit{private, personally captured} photographs of campus scenes, guaranteeing the entire dataset's absence from any pre-training corpus. This setup allows for a rigorous evaluation of true generalization, free from the confounding variable of data overlap between pre-training corpora and public benchmarks.

\begin{table}[t]
\centering
\caption{Comparison results on WebAIG-25 dataset.}
\vspace{-10pt}
\label{tab:compare-webaig25}
\begin{adjustbox}{width=0.6\linewidth} 
  \begin{tabular}{l|ccc}
  \toprule
  Method & R.Acc & F.Acc & Avg. \\
  \midrule
  CNNSpot \textsubscript{\textcolor{blue}{(CVPR'20)}}~\cite{wang2020cnn} & 97.1 & 8.3 & 52.7 \\
  UnivFD \textsubscript{\textcolor{blue}{(CVPR'23)}}~\cite{ojha2023towards} & 59.2 & 29.8 & 44.5 \\
  NPR \textsubscript{\textcolor{blue}{(ECCV'24)}}~\cite{cozzolino2024zero} & 93.5 & 34.9 & 64.2 \\
  SAFE \textsubscript{\textcolor{blue}{(KDD'25)}}~\cite{li2025improving} & 99.7 & 21.3 & 60.5 \\
  FreqNet \textsubscript{\textcolor{blue}{(AAAI'24)}}~\cite{tan2024frequency} & 67.0 & 26.0 & 46.5 \\
  C2P-CLIP \textsubscript{\textcolor{blue}{(AAAI'25)}}~\cite{tan2025c2p} & 84.7 & 31.0 & 57.9 \\
  PatchShuffle \textsubscript{\textcolor{blue}{(NeurIPS'24)}}~\cite{zheng2024breaking} & 54.9 & 45.0 & 50.0 \\
  \midrule
  \rowcolor{mygray}
  CLIP \textsubscript{(2021.2.26)}& 46.7 & 71.1 & 58.9 \\
  \rowcolor{mygray}
  Siglip \textsubscript{(2023.3.27)}& 31.3 & 44.1 & 37.7 \\
  \rowcolor{mygray}
  Siglip2 \textsubscript{(2025.2.21)}& 69.5 & 55.8 & 62.7 \\
  \rowcolor{mygray}
  \textcolor{blue}{Meta CLIP} \textsubscript{\textcolor{blue}{(2023.9.28)}}& 36.2 & 55.6 & 45.9 \\
  \rowcolor{mygray}
  \textcolor{blue}{Meta CLIP-2} \textsubscript{\textcolor{blue}{(2025.7.29)}} & 27.4 & 75.4 & 51.4 \\
  \rowcolor{mygray}
  \textcolor{blue}{PE} \textsubscript{\textcolor{blue}{(2025.4.17)}} & 44.4 & 91.7 & \underline{68.1} \\
  \rowcolor{mygray}
  \textcolor{blue}{DINOv2} \textsubscript{\textcolor{blue}{(2023.10.26)}}& 40.9 & 54.2 & 47.6 \\
  \rowcolor{mygray}
  \textcolor{blue}{DINOv3} \textsubscript{\textcolor{blue}{(2025.8.14)}} & \textbf{97.5} & \textbf{91.4} & \textbf{94.5} \\
  \bottomrule
  \end{tabular}
  \end{adjustbox}
  \vspace{-10pt}
\end{table}

The results on this verifiably unseen dataset (Table~\ref{tab:compare-webaig25}) are striking, revealing distinct and opposing biases rooted in different training paradigms. \textbf{Specialized detectors} show a strong bias towards "real", achieving high accuracy on our private real photos while catastrophically failing on novel fakes. Conversely, modern \textbf{Vision-Language Models}, trained via text-image contrastive learning, exhibit an opposite bias: they excel at identifying novel fakes but struggle with our out-of-distribution real photos, suggesting their core training objective of aligning images with text descriptions limits their ability to generalize to visual distributions, like our private photos, that fall outside this vast but specific semantic space. Amidst these widespread biases, the self-supervised \textbf{DINOv3} emerges as the sole exception, achieving exceptional performance on both real and fake images. This suggests that while data exposure dictates the biases of most models, only DINOv3's robust, fine-grained visual representations, learned without textual shortcuts, can generalize to both unseen real and synthetic distributions.


\begin{table*}[th!]
\centering

\captionof{table}{Comparison of Text–Image Similarities on In-the-Wild Dataset}
\vspace{-10pt}
\label{tab:textimgmatch}
\begin{adjustbox}{width=0.9\linewidth}
  \begin{tabular}{l|cccccccc}
  \toprule
  \multirow{3}{*}{Method} & \multicolumn{4}{c}{Chameleon}& \multicolumn{4}{c}{SocialRF}\\
  \cmidrule(r){2-5} \cmidrule(r){6-9}
   & \multicolumn{2}{c}{Top-1} & \multicolumn{2}{c}{Top-2} & \multicolumn{2}{c}{Top-1} & \multicolumn{2}{c}{Top-2} \\
   \cmidrule(r){2-3} \cmidrule(r){4-5} \cmidrule{6-7} \cmidrule{8-9}
   &Matched Text& Similarity Score&Matched Text& Similarity Score&Matched Text& Similarity Score&Matched Text& Similarity Score\\
  \midrule

    CLIP \textsubscript{(2021.2.26)} &modern design&0.628&portrait&0.345&forged&0.332&urban&0.253\\

  Siglip \textsubscript{(2023.3.27)}&unaltered&0.318&original&0.212&edited&0.232&original&0.222\\
  Siglip2 \textsubscript{(2025.2.21)}&genuine&0.385&urban&0.250&portrait&0.202&vintage&0.191\\
  \textcolor{blue}{Meta CLIP} \textsubscript{\textcolor{blue}{(2023.9.28)}}&AI generated&0.678&deepfake&0.091&AI generated&0.902&original&0.024\\
  \textcolor{blue}{Meta CLIP-2} \textsubscript{\textcolor{blue}{(2025.7.29)}} &AI generated&0.828&deepfake&0.064&AI generated&0.924&deepfake&0.038\\
  \textcolor{blue}{PE} \textsubscript{\textcolor{blue}{(2025.4.17)}}&AI generated&0.861&deepfake&0.021&AI generated&0.943&deepfake&0.031\\

  \bottomrule
  \multirow{3}{*}{Method} & \multicolumn{4}{c}{CommunityAI}& \multicolumn{4}{c}{WebAIG-25} \\
  \cmidrule(r){2-5} \cmidrule(r){6-9}
   & \multicolumn{2}{c}{Top-1} & \multicolumn{2}{c}{Top-2} & \multicolumn{2}{c}{Top-1} & \multicolumn{2}{c}{Top-2} \\
      \cmidrule(r){2-3} \cmidrule(r){4-5} \cmidrule{6-7} \cmidrule{8-9}

&Matched Text& Similarity Score&Matched Text& Similarity Score&Matched Text& Similarity Score&Matched Text& Similarity Score\\
    \midrule

  CLIP \textsubscript{(2021.2.26)}&portrait&0.346&nature&0.326&urban&0.332&midjourney\_images&0.260 \\
  Siglip \textsubscript{(2023.3.27)}&AIGIBench&0.283&real&0.281&fake&0.278&original&0.246\\
  Siglip2 \textsubscript{(2025.2.21)}&urban&0.209&portrait&0.201&urban&0.212&portrait&0.194\\
  \textcolor{blue}{Meta CLIP} \textsubscript{\textcolor{blue}{(2023.9.28)}}&AI generated&0.726&deepfake&0.086&midjourney\_images&0.604&AI generated&0.284\\
  \textcolor{blue}{Meta CLIP-2} \textsubscript{\textcolor{blue}{(2025.7.29)}}&AI generated&0.858&deepfake&0.041&midjourney\_images&0.621&AI generated&0.308\\
  \textcolor{blue}{PE} \textsubscript{\textcolor{blue}{(2025.4.17)}}&AI generated&0.878&CommunityAI&0.029&midjourney\_images&0.722&AI generated&0.218\\

  \bottomrule
   
  \end{tabular}
  \end{adjustbox}
  \vspace{-10pt}
\end{table*}

\vspace{-3mm}
\subsection{Probing VFM's Latent Alignment with Forgery Concepts}
\vspace{-1mm}

To investigate the mechanism behind the success of Vision-Language Models (VLMs), we hypothesize that their pre-training exposure has allowed them to learn a powerful semantic shortcut: associating synthetic images with high-level, forgery-related textual concepts. This provides a potential high-level cue for detection that we can directly probe.

To test this specific hypothesis, we designed a text-image similarity probing experiment applicable \textbf{only to VLM-based backbones}, as it relies on their inherent text-encoder capabilities. This analysis is therefore not applicable to self-supervised models like \textbf{DINOv3}, whose remarkable performance must stem from a different, purely visual mechanism. First, we constructed a comprehensive text pool categorized into three groups:

\begin{compactitem}
    \item \textbf{Forgery-Related Concepts:} \emph{[`fake', `real', `AI generated', `authentic', `manipulated', `synthetic', etc.]}
    \item \textbf{Content-Related Concepts:} \emph{[`sunset', `landscape', `portrait', `abstract art', `technology', `nature', etc.]}
    \item \textbf{Source-Related Concepts:} \emph{[`GenImage', `ADM', `BigGAN', `glide', `Midjourney', etc.]}
\end{compactitem}

For the experiment, we randomly sampled 2,000 \textbf{fake images} from each in-the-wild dataset. We then computed the cosine similarity between each image's visual embedding and the textual embeddings of all concepts in our pool. The top-matching text concepts for each VLM are reported to reveal their latent semantic alignments.


Table~\ref{tab:textimgmatch} reveals a significant generational divide in semantic understanding. Older models like \textbf{CLIP} and \textbf{Siglip} consistently fail to associate fake images with forgery-related concepts, instead matching them to generic terms like `portrait' or `urban'. Their latent space essentially lacks a concept of "fakeness." In stark contrast, modern VLMs like \textbf{Meta CLIP-2} and \textbf{PE} demonstrate a remarkably strong and consistent alignment. Across both established in-the-wild datasets and our new \textbf{WebAIG-25}, their top-matched concepts for fake images are overwhelmingly forgery-related terms like "AI generated" or "midjourney\_images". It confirms that modern VLMs excel not just by recognizing visual artifacts, but by understanding the very concept of a "fake" image.

\vspace{-3mm}
\subsection{Visualization}
\vspace{-1mm}

\begin{figure}[th!]
  \centering
  \includegraphics[width=0.9\linewidth]{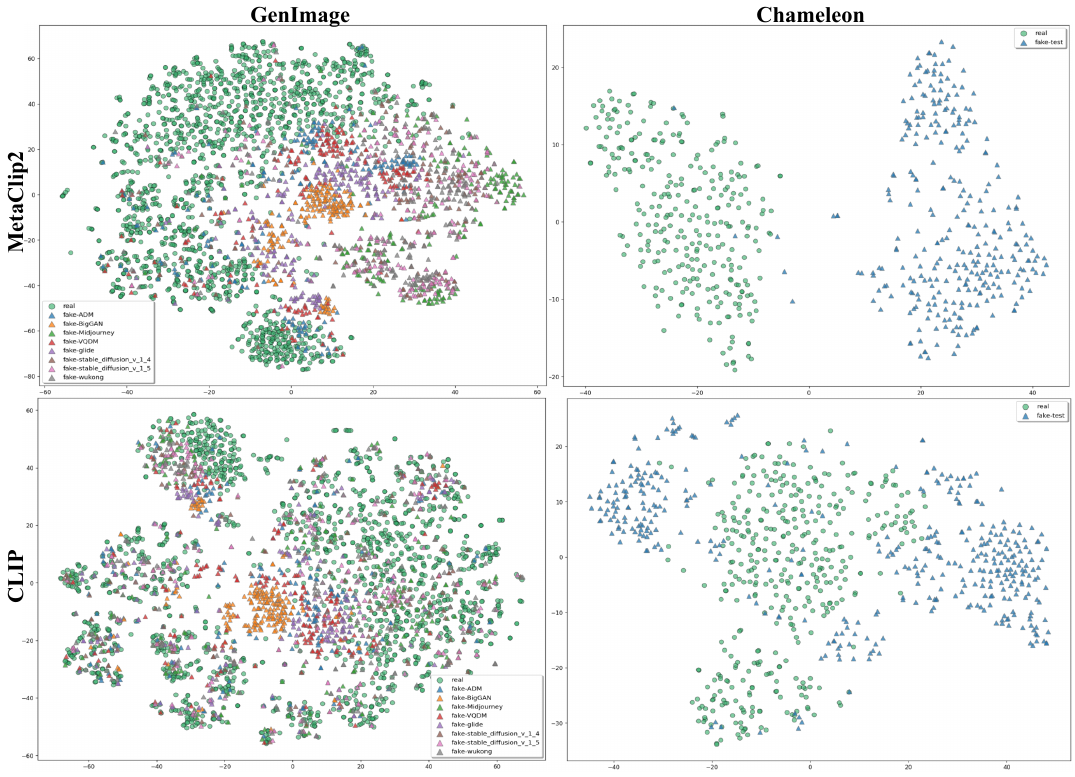}
  \vspace{-10pt}
  \caption{T-SNE Visualization of CLIP and Meta CLIP-2 on GenImage and Chameleon.}
  \label{fig:tsne}
  \vspace{-15pt}
\end{figure}

The t-SNE visualizations in Fig.~\ref{fig:tsne} offer a compelling qualitative explanation for the quantitative performance gaps in Tab.~\ref{tab:compare-genimagechameleon}. On the curated GenImage dataset, both Meta CLIP-2 and CLIP exhibit similarly entangled feature spaces, which aligns with the comparable, high performance of both our VFM baselines (89.5\%) and CLIP-based detectors like \textbf{UnivFD} (88.8\%). \textbf{However, this parity vanishes on the in-the-wild Chameleon dataset.} There, the performance of \textbf{UnivFD} collapses to 57.2\%, a failure visually mirrored in its CLIP backbone's chaotic and inseparable feature space. In stark contrast, \textbf{Meta CLIP-2}'s near-perfect accuracy (91.8\%) is directly explained by its t-SNE plot, which reveals a remarkably clean separation between real and fake clusters. This direct link between feature separability and downstream performance confirms that Meta CLIP-2's advantage stems from a fundamentally more discriminative representation for in-the-wild content, acquired through pre-training exposure (Fig.~\ref{fig:tsne}).
\vspace{-1mm}

\vspace{-2mm}
\section{Conclusion}
\vspace{-2mm}
This paper demonstrated that for in-the-wild AI-generated image detection, a simple linear classifier built upon a modern Vision Foundation Model (VFM) decisively outperforms specialized forensic detectors, akin to bringing a gun to a knife fight. We attribute this stark performance gap not to architectural novelty but to the VFM's pre-training on vast, contemporary web data that includes synthetic media. Our findings have two critical implications: First, for practical forensics, leveraging the evolving representations of the latest foundation models is a more robust strategy than designing static detectors. Second, for academic research, a stricter evaluation protocol that uses test data guaranteed to be novel to a model's entire pre-training corpus is paramount to truly assess generalization.

\bibliographystyle{ieeetr}
\small
\bibliography{refs}


\end{document}